\pdfoutput=1

\documentclass[11pt]{article}

\usepackage{EMNLP2022}

\usepackage{times}
\usepackage{latexsym}

\usepackage[T1]{fontenc}

\usepackage[utf8]{inputenc}

\usepackage{microtype}

\usepackage{inconsolata}

\usepackage{amsmath}
\usepackage{graphicx}
\usepackage{booktabs}
\usepackage{arydshln}
\usepackage{tabularx}
\usepackage{amsfonts}
\usepackage{multirow}
\usepackage{multicol}
\usepackage{colortbl}
\usepackage{hyperref}
\usepackage{cleveref}
\usepackage{xspace}
\usepackage{physics}

\newcommand{\yijia}[1]{}  

\crefformat{section}{\S#2#1#3} 
\crefformat{subsection}{\S#2#1#3}
\crefformat{subsubsection}{\S#2#1#3}
\newcommand{\refequ}[1]{Equation~(\ref{#1})}
\newcommand{\reffig}[1]{Figure~\ref{#1}}
\newcommand{\reftab}[1]{Table~\ref{#1}}

\def\eg{\textit{e.g.}\xspace}
\def\Eg{\textit{E.g.}\xspace}

\def\etc{\textit{etc.}\xspace}
\def\ie{\textit{i.e.}\xspace}

\definecolor{myblue}{RGB}{0, 112, 192}

\makeatletter
\newcommand{\printfnsymbol}[1]{%
  \textsuperscript{\@fnsymbol{#1}}%
}
\makeatother

\title{FormLM: Recommending Creation Ideas for Online Forms by Modelling~Semantic~and~Structural~Information}

\author{
Yijia Shao\textsuperscript{\rm 1}\thanks{\indent The contributions by Yijia Shao, Yifan Zhong and Hongwei Han have been conducted and completed during their internships at Microsoft Research Asia, Beijing, China.} \hspace{0.5em}
Mengyu Zhou\textsuperscript{\rm 2}\thanks{\indent Corresponding author.}\hspace{0.5em}
Yifan Zhong\textsuperscript{\rm 3}\printfnsymbol{1}\hspace{0.5em}
Tao Wu\textsuperscript{\rm 4}\hspace{0.5em}
Hongwei Han\textsuperscript{\rm 5}\printfnsymbol{1}\hspace{0.5em}\\
\textbf{Shi Han}\textsuperscript{\rm 2}\hspace{0.5em}
\textbf{Gideon Huang}\textsuperscript{\rm 4}\hspace{0.5em}
\textbf{Dongmei Zhang}\textsuperscript{\rm 2}\hspace{0.5em}\\
\textsuperscript{\rm 1}Peking University
\textsuperscript{\rm 2}Microsoft Research
\textsuperscript{\rm 3}Fudan University
\textsuperscript{\rm 4}Microsoft
\textsuperscript{\rm 5}Tsinghua University\\
\texttt{shaoyj@pku.edu.cn},
\texttt{\{\href{mailto:mezho@microsoft.com}{mezho}, twu, shihan, gihuang, dongmeiz\}@microsoft.com}\\
\texttt{yfzhong20@fudan.edu.cn},
\texttt{hhw20@mails.tsinghua.edu.cn}
}

\begin{document}
\maketitle
\begin{abstract}
\label{sec:abstract}
Online forms are widely used to collect data from human and have a multi-billion market. Many software products provide online services for creating semi-structured forms where questions and descriptions are organized by predefined structures. However, the design and creation process of forms is still tedious and requires expert knowledge. To assist form designers, in this work we present \textbf{FormLM} to model online forms (by enhancing pre-trained language model with form structural information) and recommend form creation ideas (including question / options recommendations and block type suggestion). For model training and evaluation, we collect the first public online form dataset with 62K online forms. Experiment results show that FormLM significantly outperforms general-purpose language models on all tasks, with an improvement by 4.71 on Question Recommendation and 10.6 on Block Type Suggestion in terms of ROUGE-1 and Macro-F1, respectively.

\end{abstract}

\section{Introduction}
\label{sec:intro}

\begin{figure}[th]
    \resizebox{\columnwidth}{!}{%
    \centering 
    \includegraphics{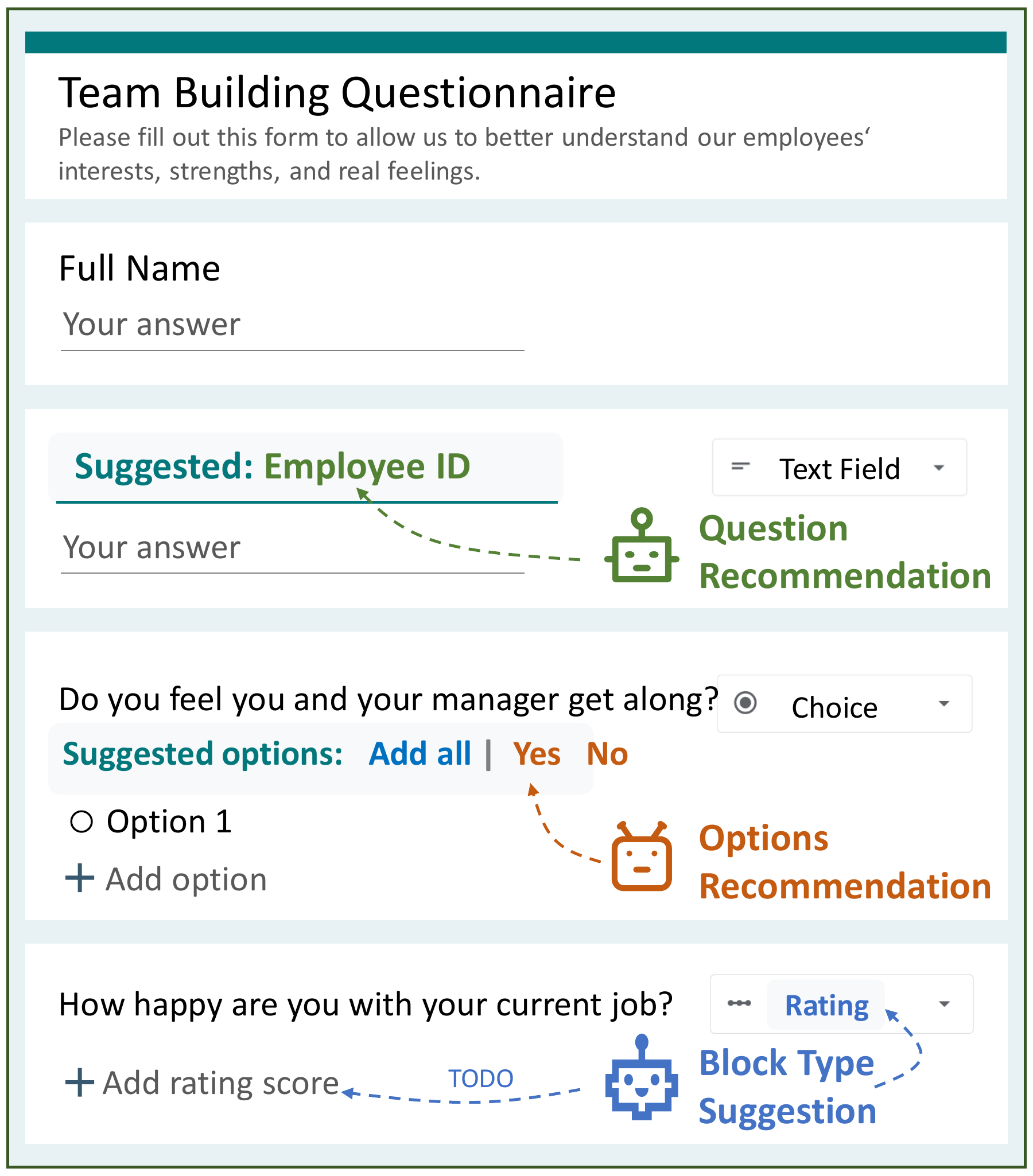}
    }
    \caption{An Example Online Form with the Three Tasks of Intelligent Form Creation Ideas.}
    \label{Fig.demo}
\end{figure}

\yijia{Introduce online forms.}
Online forms are widely used to collect data in everyday scenarios such as feedback gathering~\citep{ilieva2002online}, application system~\citep{sylva2009recruitment}, research surveys~\citep{yarmak2017online}, \etc With a multi-billion market~\citep{market}, 
many software products -- such as Survey Monkey~\citep{abd2018review}, Google~\citep{mondal2018using} and Microsoft Forms~\citep{rhodes2019creating} -- provide services to help users create online forms which consist of multiple blocks (\eg, \reffig{Fig.demo}).

\yijia{Introduce what we want to achieve by researching online forms.}

However, there are obstacles preventing the creation of well-designed online forms, which could hurt response rate and quality~\citep{krosnick2018questionnaire}. For each form question, form designers need to write an informative title, specify its type, and provide other required components. This process is tedious and time-consuming even for experienced users. Also, non-experts may be unsure about what question to add or which question type to choose. To improve the experience and efficiency of form composing, it is desirable that online form services could recommend creation ideas to form designers.

To address the above demands, in \cref{sec:problem} we identify three machine learning (ML) tasks of \textbf{Form Creation Ideas}, including \textit{Question Recommendation}, \textit{Block Type Suggestion}, and \textit{Options Recommendation}. For example, in \reffig{Fig.demo}, when one adds a text field block as the second block, the Question Recommendation suggests ``Employee ID'' for the question based on the existing content (form title, description, and the first question ``Full Name''). When editing the third choice question block, the Options Recommendation suggests ``Yes'' and ``No'' as candidate options. Finally, if the user types ``How happy are you with your current job?'' for the fourth block but hasn't selected a block type yet, the Block Type Suggestion predicts it as a rating type block.

The above tasks require a specifically designed model to understand semi-structured forms, where natural language (NL) text is organized by predefined structures. 
A form is composed of a title, a description, and a series of blocks. For each block, its subcomponents also follow unique structures. For example, a \textit{Choice} block contains a list of options which serve as candidate answers to the question displayed in the block title. 
Existing pre-trained language models (PLMs) focus on general-purpose free-form NL text~\citep{devlin-etal-2019-bert,yang2019xlnet}. They may provide a good starting point to model the rich semantic information within NL contents of a form. However, they cannot directly handle the extra structural information of the form. 
\textit{Is it possible to infuse a PLM with structural information of online forms?}

\yijia{Introduction of our approach.}
In this paper, we propose \textbf{FormLM} to model both the semantic and structural information of online forms. As we will discuss in \cref{sec:method}, there are three key parts of FormLM. 
First, the form serialization procedure, which represents a form as a tree and converts it into a token sequence without information loss. 
Second, inheriting existing PLM with a small number of additional parameters: FormLM inherits the parameters of BART~\citep{lewis-etal-2020-bart} to leverage its language modelling capabilities. Also, by adding extra biases to the attention layers, FormLM explicitly handles the structural information. 
Third, continual pre-training with collected online forms: for better downstream application: 
We propose two structure-aware objectives -- Span Masked Language Model and Block Title Permutation -- to continually pre-train FormLM on top of the inherited and additional parameters.

\yijia{Our model's performance.}
We evaluate FormLM on Form Creation Ideas tasks using our \textbf{OOF} (\textbf{O}pen \textbf{O}nline \textbf{F}orms) dataset. This dataset (see \cref{sec:dataset}) is created by crawling and parsing public forms on the Web. Comparing to PLMs such as BART, FormLM improves the ROUGE-1 score from 32.82 to 37.53 on Question Recommendation, and the Macro-F1 score from 73.3 to 83.9 on Block Type Suggestion.


In summary, our main contributions are:
\begin{itemize}
    \item We put forward the problem of online form modeling and formally define a group of tasks on Form Creation Ideas. To the best of our knowledge, these problems have not been systematically studied before.

    
    \item FormLM is proposed by us to model both the semantic and structural information by enhancing PLM with form serialization, structural attention and continual pre-training.
    
    
    \item The public OOF dataset with 62k forms is constructed by us. To the best of our knowledge, this is the first public online form dataset. OOF dataset, FormLM code and models are also open sourced at \url{https://github.com/microsoft/FormLM}.
    
    \item Comprehensive experiments -- especially baseline comparisons, ablation studies, design choices and empirical studies -- are designed and run by us to evaluate the effectiveness of FormLM on the tasks of Form Creation Ideas with the form dataset.
\end{itemize}

\section{Preliminaries}
\label{sec:preliminaries}
In this section, we further elaborate the predefined structure in online forms, and introduce our collected dataset.

\subsection{Online Form Structure}
\label{sec:form structure}
Modern online form services usually allow users to create a form by piling up different types of blocks. There are eight common block types: \textit{Text Field, Choice, Time, Date, Likert, Rating, Upload}, and \textit{Description}. Each block type has a predefined structure (\eg, the options of a choice block) and corresponds to a specific layout shown in the user interface (\eg, bullet points or checkboxes of the options). 
The order of the blocks in a form usually matters because they are designed to organize questions in an easy-to-understand way, and to collect data from various related aspects. For example, in \reffig{Fig.demo}, easier profile / fact questions are asked before the preference / opinion questions.

As shown at the top of \reffig{Fig.scope}, 
an online form can be viewed as an ordered tree. The root node $T$ represents the form title, and its children nodes $\operatorname{Ch}(T)=(\text{Desc}, B_1, ...,B_N)$ represent the form description and a series of blocks. The subtree structure of $B_i$ depends on its type. For \textit{Choice} and \textit{Rating} blocks, $\operatorname{Ch}(B_i)=(\text{Type}_i, \text{Title}_i, \text{Desc}_i, C_i^{(1)}, ..., C_{i}^{(n_i)})$ where $C_i^{(k)}$ are the options or scores; For \textit{Likert}~\citep{johns2010likert} blocks,  $\operatorname{Ch}(B_i)=(\text{Type}_i, \text{Title}_i, \text{Desc}_i, R_i^{(1)}, ..., R_i^{(m_i)}, C_i^{(1)}, ..., C_i^{(n_i)})$ where $R_i^{(j)}$ are rows and $C_i^{(k)}$ are columns; For the remaining block types,  $\operatorname{Ch}(B_i)=(\text{Type}_i, \text{Title}_i, \text{Desc}_i)$. All description parts ($\text{Desc}$) are optional.


\begin{figure}[ht]
    \centering 
    \includegraphics[width=1\columnwidth]{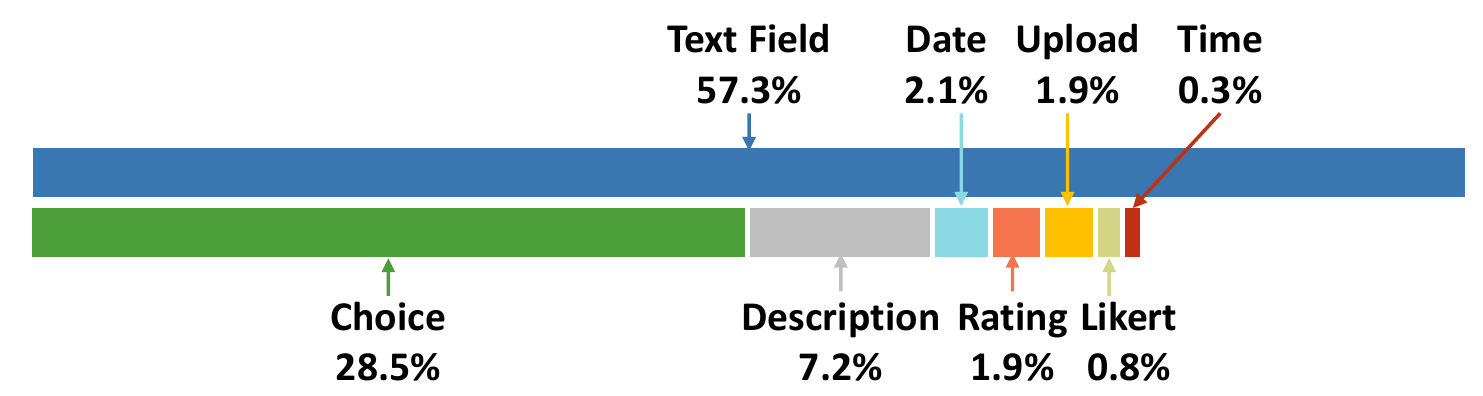}
    \caption{Distribution of Block Types in Online Forms.}
    \label{Fig.type}
\end{figure}
\subsection{Online Form Dataset}
\label{sec:dataset}

Since there is no existing dataset for online forms, we construct our own OOF (Open Online Forms) dataset by crawling public online forms created on a popular online form website. We filter out forms with low quality and only consider English forms in this work. In total, 62K public forms are collected across different domains, \eg, education, finance, medical, community activities, \etc

Due to the semi-structured nature of online forms, we further parsed the crawled HTML pages into JSON format by extracting valid contents and associating each block with its type. 
\reffig{Fig.type} shows the distribution of block types in our collected dataset. 
More details of the dataset construction and its statistics can be found in Appendix~\ref{appendix:dataset}.


\section{Form Creation Ideas}
\label{sec:creation aids}
\label{sec:problem}

As illustrated in Figure~\ref{Fig.demo}, when adding a new block, one needs to specify its type and title in the first step. Then, other required components -- such as a list of options for a \textit{Choice} block -- are added according to the block type. In this paper, we focus on the following three tasks which provide Form Creation Ideas to users in the first and later steps.

\noindent
\textbf{Question Recommendation}\quad
The Question Recommendation aims at providing users with a recommended question based on the selected block type and the previous context. Formally, the model needs to predict $\text{Title}_i$ based on $T$, $\text{Desc}$, $B_1, ..., B_{i-1}$ and $\text{Type}_i$. For example, in \reffig{Fig.demo}, it is desirable that the model could recommend ``Employee ID'' when the form designer creates a \textit{Text Field} block after the first block.

\noindent
\textbf{Block Type Suggestion}
Different from the scenario of Question Recommendation, sometimes form designers may first come up with a block title without clearly specifying its block type. The Block Type Suggestion helps users select a suitable type in this situation.
For example, for the last block of \reffig{Fig.demo}, the model will predict it as a \textit{Rating} block and suggest adding candidate rating scores if the form designer has not appointed the block type himself / herself. 
Formally, given $\text{Title}_i$ and the available context ($T, \text{Desc}, B_1, ..., B_{i-1}$), the model should predict $\text{Type}_i$ in this task.

\noindent
\textbf{Options Recommendation}
As \reffig{Fig.type} shows, \textit{Choice} blocks are frequently used in online forms. When creating a \textit{Choice} block, one should additionally provide a set of options, and the Options Recommendation helps in this case. Given the previous context ($T, \text{Desc}, B_1, ..., B_{i-1}$) and $\text{Title}_i$, the model predicts $C_i^{(1)},...,C_i^{(n_i)}$ if $\text{Type}_i=\textit{Choice}$. In this work, we expect the model to recommend a set of possible options at the same time, so the desired output of this task is $C_i^{(1)},...,C_i^{(n_i)}$ concatenated with a vertical bar. For example, in \reffig{Fig.demo}, the model may output ``Yes | No'' to recommend options for the third block.

\begin{figure*}[t]
    \centering 
    \includegraphics[width=1\textwidth]{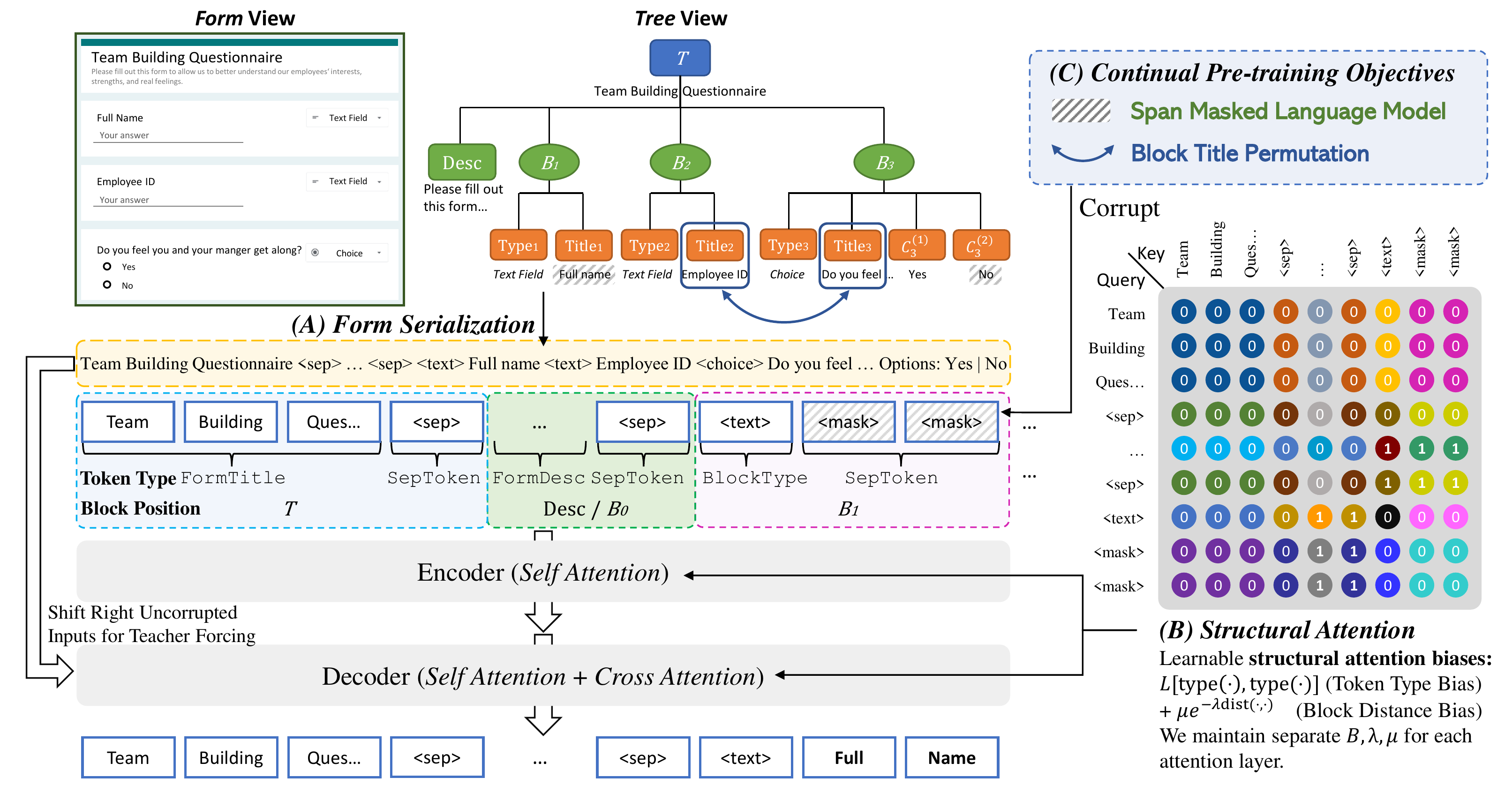}
    \caption{The Overview of FormLM Methodology. \textbf{(A)} \textbf{Form Serialization} (\cref{sec:linearization}) serializes an online form by adding block type tokens and separate tokens to preserve the tree structure. \textbf{(B)} \textbf{Structural Attention} (\cref{sec:attention}) encodes the token type and block-level distance by adding structural biases to each attention layer. Different colors in the attention bias matrix denote different items in the lookup table 
    and the number inside each circle represents the block-level distance of a token pair. \textbf{(C)} \textbf{Continual Pre-training} (\cref{sec:pre-training}) requires the model to recover the input sequence corrupted by SpanMLM and BTP. We use the cross-entropy loss between the decoder's output and the uncorrupted sequence for model optimization.}
    \label{Fig.scope}\label{fig:FormLM}
\end{figure*}

\section{Methodology}
\label{sec:method}

As discussed in \cref{sec:intro}, we propose FormLM to model forms for creation ideas. We select BART as the backbone model of FormLM because it is widely used in NL-related tasks and supports both generation and classification tasks. In the rest of this section, we will describe the design and training details of FormLM as demonstrated in \reffig{fig:FormLM}.

\subsection{Form Serialization}
\label{sec:linearization}
As discussed in \cref{sec:form structure}, an online form could be viewed as an ordered tree. In FormLM we serialize the tree into a token sequence which is compatible with the input format of common PLMs. 
\reffig{fig:FormLM}(A) depicts the serialization process which utilizes special tokens and separators. 
First, a special token is introduced for each block type to explicitly encode $\text{Type}_i$. 
Second, the vertical bar ``|'' is used to concatenate a list of related items within a block -- options / scores $C_i^{(k)}$ of a \textit{Choice} / \textit{Rating} block, and rows $R_i^{(j)}$ or columns $C_i^{(k)}$ of a \textit{Likert} block. 
Finally, multiple subcomponents of $B_i$ are concatenated using \texttt{<sep>}. 
Note that there is no information loss in the serialization process, \ie, the hierarchical tree structure of an online form can be reconstructed from the flattened sequence.

\subsection{Structural Attention}
\label{sec:attention}
Beyond adding structural information into the input sequence, in FormLM we further enhance its backbone PLM with specially designed \textit{Structural Attention} (StructAttn). 
Our intuition is that the attention calculation among tokens should consider their different roles and locations in a form. \Eg, tokens within a question title seldom correlates with the tokens of an option from another question; tokens in nearby blocks (or even the same block) are usually stronger correlated with each other than those from distant blocks. 

As illustrated in \reffig{Fig.scope}(B), StructAttn encodes the structural information of an online form by adding two bias terms based on the token type (\ie, the role that a token plays in the flattened sequence) and the block-level position. For each attention head, given the query matrix $\vb{Q}=[\vb{q_1}, \cdots, \vb{q_n}]^\top\in\mathbb{R}^{n\times d_k}$, the key matrix $\vb{K}=[\vb{k_1}, \cdots, \vb{k_m}]^\top\in\mathbb{R}^{m\times d_k}$, and the value matrix $\vb{V}=[\vb{v_1}, \cdots, \vb{v_m}]^\top\in\mathbb{R}^{m\times d_v}$, the original output is calculated by 
\begin{equation}
    \hat{\vb{A}} = \frac{\vb{Q} \vb{K}^{\top}}{\sqrt{d_{k}}},
    \operatorname{Attn}(H)=\operatorname{softmax}(\hat{\vb{A}})\vb{V}
\end{equation}

In FormLM, we add two biases to $\hat{A}$ and the attention head output of StructAttn is calculated by
\begin{equation}
\resizebox{\columnwidth}{!}{$%
\begin{gathered}
\vb{A}_{ij} = \hat{\vb{A}}_{ij} + L[\operatorname{type}(\vb{q_i}),\operatorname{type}(\vb{k_j})] + \mu e^{-\lambda \operatorname{d}(\vb{q_i},\vb{k_j})} \\
    \operatorname{Attn}(H)=\operatorname{softmax}(\vb{A})\vb{V}
\end{gathered}$%
}
\label{eq:structAttn}
\end{equation}

In \refequ{eq:structAttn}, the token type bias is calculated based on a learnable lookup table $L[\cdot,\cdot]$ in each attention layer, and the lookup key $\operatorname{type}(\cdot)$ is the type of the corresponding token within the form structure. Specifically, in our work, $\operatorname{type}(\cdot)$ is chosen from 9 token types: \texttt{FormTitle}, \texttt{FormDesc}, \texttt{BlockTitle}, \texttt{BlockDesc}, \texttt{Option}, \texttt{LikertRow}, \texttt{LikertColumn}, \texttt{BlockType}, \texttt{SepToken}. If $\vb{Q}$ or $\vb{K}$ corresponds to the flattened sequence given by form serialization, $\operatorname{type}(\cdot)$ can be directly obtained from the original form tree; otherwise, in generation tasks, $\vb{Q}$ or $\vb{K}$ may correspond to the target, and we set $\operatorname{type}(\cdot)$ as the expected output token type, \ie, \texttt{BlockTitle} when generating the question and \texttt{Option} when generating the options.

Another bias term in \refequ{eq:structAttn} is calculated by an exponential decay function to model the relative block-level position, where $\operatorname{d}(\vb{q_i}, \vb{k_j})$ is the block-level distance between the corresponding tokens of $\vb{q_i}$ and $\vb{k_j}$ on the form tree. To make $\operatorname{d}(\vb{q_i}, \vb{k_j})$ well-defined for each token pair, we set $\text{Desc}$ as the 0-th block ($B_0$) and specify $\operatorname{d}(\vb{q_i}, \vb{k_j})$ as 0 if $\operatorname{type}(\vb{q_i})$ or $\operatorname{type}(\vb{k_j})$ is equal to \texttt{FormTitle}. Note that there are two parameters $\lambda,\mu$ in this term. We make them trainable and constrain their values to be positive to ensure tokens in neighboring blocks give more attention to each other.

We apply StructAttn to three parts of FormLM, self attentions of FormLM encoder, self attentions and cross attentions of FormLM decoder. $\vb{Q}, \vb{K}, \vb{V}$ of encoder self attentions and $\vb{K}, \vb{V}$ of decoder cross attentions correspond to the source sequence; while $\vb{Q}, \vb{K}, \vb{V}$ of decoder self attentions and $\vb{Q}$ of decoder cross attentions correspond to the target sequence.
In classification, both the source and the target are the flattened form; while in generation, the target is the recommended question or options. 

In \cref{sec:ablation}, we will prove the effectiveness of StructAttn through ablation studies and comparing alternative design choices of StructAttn.

\subsection{Continual Pre-training}
\label{sec:pre-training}
Note that it is difficult to train a model for online forms from scratch due to the limited data. To effectively adapt FormLM to online forms, we conduct continual pre-training on the training set of our collected dataset (see \cref{sec:dataset}) with the following two structure-aware objectives.

\noindent
\textbf{Span Masked Language Model (SpanMLM)}\quad
We adapt the masked language model (MLM) to forms by randomly selecting and masking some nodes on the form tree within the masking budget. Compared to SpanBERT~\citep{joshi-etal-2020-spanbert} which improves the MLM objective by masking a sequence of complete words, we do the masking in a higher level of granularity based on the form structure. Our technique masks a block title, option, \etc, instead of arbitrarily masking subword tokens. The latter was proven suboptimal in~\citet{joshi-etal-2020-spanbert, zhang-etal-2019-ernie}. 
Specifically, we use a masking budget of 15\% and replacing 80\% of the masked tokens with \texttt{<MASK>}, 10\% with random tokens and 10\% with the original tokens.

\noindent
\textbf{Block Title Permutation (BTP)}\quad
As discussed in \cref{sec:form structure}, each block can be viewed as a subtree. 
We introduce the block title permutation objective by permuting block titles in a form and requiring the model to recover the original sequence with the intuition that the model needs to understand the semantic relationship between $B_i$ and $\operatorname{Ch}(B_i)$ to solve this challenge. We randomly shuffle all the block titles to construct the corrupted sequence.

Following the pre-training process of BART, we unify these two objectives by optimizing a reconstruction loss, \ie, we input the sequence corrupted by SpanMLM and BTP and optimize the cross-entropy loss between the decoder's output and the original intact sequence.


\section{Experiments}
\label{sec:experiments}

\begin{table*}[t]
\centering
\resizebox{\textwidth}{!}{%
\setlength{\extrarowheight}{0pt}
\addtolength{\extrarowheight}{\aboverulesep}
\addtolength{\extrarowheight}{\belowrulesep}
\setlength{\aboverulesep}{0pt}
\setlength{\belowrulesep}{0pt}
\begin{tabular}{lcccccccc} 
\toprule
                                              & \multicolumn{3}{c}{\textbf{Question Recommendation}} & \multicolumn{3}{c}{\textbf{Options Recommendation}} & \multicolumn{2}{c}{\textbf{Block Type Suggestion}}  \\
                                              & R1             & R2             & RL                      & R1             & R2             & RL                & Macro-F1            & Accuracy                                \\ 
\hline
RoBERTa                                       & ~-             & -              & -                       & -              & -              & -                 & 73.7\small{$\pm$0.02}          & 85.8\small{$\pm$0.46}                                \\
GPT-2      &     22.82\small{$\pm$0.22}  &  9.71\small{$\pm$0.04}     &  22.37\small{$\pm$0.20}              &       17.84\small{$\pm$0.10}              &   11.38\small{$\pm$0.05}           &     16.94\small{$\pm$0.10}              &   74.2\small{$\pm$0.16}    &  85.6\small{$\pm$0.06}                                   \\
MarkupLM                                      & -              & -              & -                       & -              & -              & -                 & 79.8\small{$\pm$0.27}          & 88.6\small{$\pm$0.13}                                \\
$\text{BART}_\text{BASE}$                                     & 31.48\small{$\pm$0.16}          & 15.89\small{$\pm$0.18}          & 30.91\small{$\pm$0.16}                   & 43.53\small{$\pm$0.32}          & 31.81\small{$\pm$0.21}          & 41.5\small{$\pm$0.29}              & 73.4\small{$\pm$0.31}          & 85.6\small{$\pm$0.17}                                \\
BART                                          & 32.82\small{$\pm$0.05}          & 17.06\small{$\pm$0.20}          & 32.18\small{$\pm$0.05}                   & 46.12\small{$\pm$0.12}          & 33.74\small{$\pm$0.08}          & 43.85\small{$\pm$0.12}             & 73.3\small{$\pm$0.28}          & 85.3\small{$\pm$0.08}                                \\ 
\hline
$\text{FormLM}_\text{BASE}$                                   & 35.9\small{$\pm$0.08}           & 18.27\small{$\pm$0.10}          & 35.23\small{$\pm$0.04}                   & 44.14\small{$\pm$0.06}          & 32.39\small{$\pm$0.16}          & 42.21\small{$\pm$0.10}             & 83.0\small{$\pm$0.06}            & 90.7\small{$\pm$0.09}                                \\
\rowcolor[rgb]{0.925,0.925,0.925}~~~~$\uparrow$~~$\text{BART}_\text{BASE}$ & 4.42           & 2.38           & 4.32                    & 0.61           & 0.58           & 0.71              & 9.6           & 5.1                                 \\
FormLM                                        & \textbf{37.53}\small{$\pm$0.07} & \textbf{19.70}\small{$\pm$0.15} & \textbf{36.78}\small{$\pm$0.12}          & \textbf{47.24}\small{$\pm$0.02} & \textbf{34.65}\small{$\pm$0.14} & \textbf{44.91}\small{$\pm$0.08}    & \textbf{83.9}\small{$\pm$0.11} & \textbf{91.0}\small{$\pm$0.08}                       \\
\rowcolor[rgb]{0.925,0.925,0.925}~~~~$\uparrow$~~BART     & 4.71           & 2.64           & 4.6                     & 1.12           & 0.91           & 1.06              & 10.6          & 5.7                                 \\
\bottomrule
\end{tabular}
}
\caption{Results of FormLM and the Baseline Models on the Tasks of Form Creation Ideas. Note that RoBERTa and MarkupLM are encoder-only models, thus cannot be directly applied to generation tasks. We leave their results blank for Question and Options Recommendations where ROUGE scores (R1, R2, RL) are used to evaluate these two generation tasks. Both the averaged metric and its standard deviation (as subscript) are reported for each result over 3 runs. The two gray rows (with up arrow $\uparrow$) show the improvement of FormLM over its backbone model.}
\label{table:mainResults}
\end{table*}


\subsection{Evaluation Data and Metrics}
\label{sec:data_and_metric}
We evaluate FormLM and other models on the three tasks of Form Creation Ideas (\cref{sec:creation aids}) with our OOF dataset (\cref{sec:dataset}). The 62k public forms are split into 49,904 for training, 6,238 for validation, and 6,238 for testing. For each task, random sampling is further performed to construct an experiment dataset. 
Specifically, for each task, we randomly select no more than 5 samples from a single form to avoid sample bias introduced by those lengthy forms.
For Question Recommendation and Block Type Suggestion, each sample corresponds to a block and its previous context (see \cref{sec:problem}). 239,544, 29,558 and 29,466 samples are selected for training, validation and testing, respectively. For Options Recommendation, each sample corresponds to a \textit{Choice} block with context. 124,994, 15,640 and 15,867 samples are selected for training, validation, and testing.

For Question and Options Recommendations, following the common practice in natural language generation research, we adopt ROUGE\footnote{We use the Hugging Face implementation to calculate the ROUGE score, \url{https://huggingface.co/metrics/rouge}.}~\citep{lin-2004-rouge} scores with the questions/options composed by human as the ground truth. During option recommendation, because the model is expected to recommend a list of options at once, we concatenate options with a vertical bar (described in \cref{sec:linearization}) 
for the comparison of generated results and ground truths. Since it is difficult to have a thorough evaluation of the recommendation quality through the automatic metric, we further include a qualitative study in Appendix~\ref{appendix:study} and conduct human evaluations for these two generation tasks (details in Appendix~\ref{sec:human}). For Block Type Suggestion, both accuracy and Macro-F1 are reported to take account of the class imbalance issue.

\subsection{Baselines}
\label{sec:baselines}
As there was no existing system or model specifically designed for forms, we compare FormLM with three general-purposed PLMs -- RoBERTa~\citep{liu2020roberta}, GPT-2~\citep{radford2019language} and BART~\citep{lewis-etal-2020-bart}, which represent widely-used encoder, decoder, encoder-decoder based models, respectively. To construct inputs for these PLMs, we concatenate NL sentences in the available context (see \cref{sec:creation aids}).

MarkupLM~\citep{li-etal-2022-markuplm}, a recent model for web page modeling, is also chosen as a baseline since forms can be displayed as HTML pages on the Internet. To keep accordance with the original inputs of MakupLM, we remove the tags without NL text (\eg, \texttt{<script>}, \texttt{<style>}) in the HTML file in OOF dataset.

The number of parameters of each model can be found in Appendix~\ref{sec:config}.


\subsection{FormLM Implementation}
\label{sec:setups}
We implement FormLM using the Transformers library~\citep{wolf-etal-2020-transformers}. FormLM and $\text{FormLM}_{\text{BASE}}$ are based on the architecture and parameters of BART\footnote{\url{https://huggingface.co/facebook/bart-large}} and $\text{BART}_\text{BASE}$\footnote{\url{https://huggingface.co/facebook/bart-base}} respectively.

For continual pre-training, we train FormLM for 15k steps on 8 NVIDIA V100 GPUs with the total batch size of 32 using the training set of the OOF dataset. For all the three tasks of Forms Creation Ideas, we fine-tune FormLM and all baseline models for 5 epochs with the total batch size of 32 and the learning rate of 5e-5. 
More pre-training and fine-tuning details are described in Appendix~\ref{appendix:implementation}.

In the rest of this paper, each experiment with randomness is run for 3 times and reported with averaged evaluation metrics.

\subsection{Main Results}
\label{sec: main results}
For FormLM and the baseline models (see \cref{sec:baselines}), \reftab{table:mainResults} shows the results on the Form Creation Ideas tasks. 
FormLM significantly outperforms the baselines on all tasks. 

\begin{figure}[t]
    \centering 
    \includegraphics[width=\columnwidth]{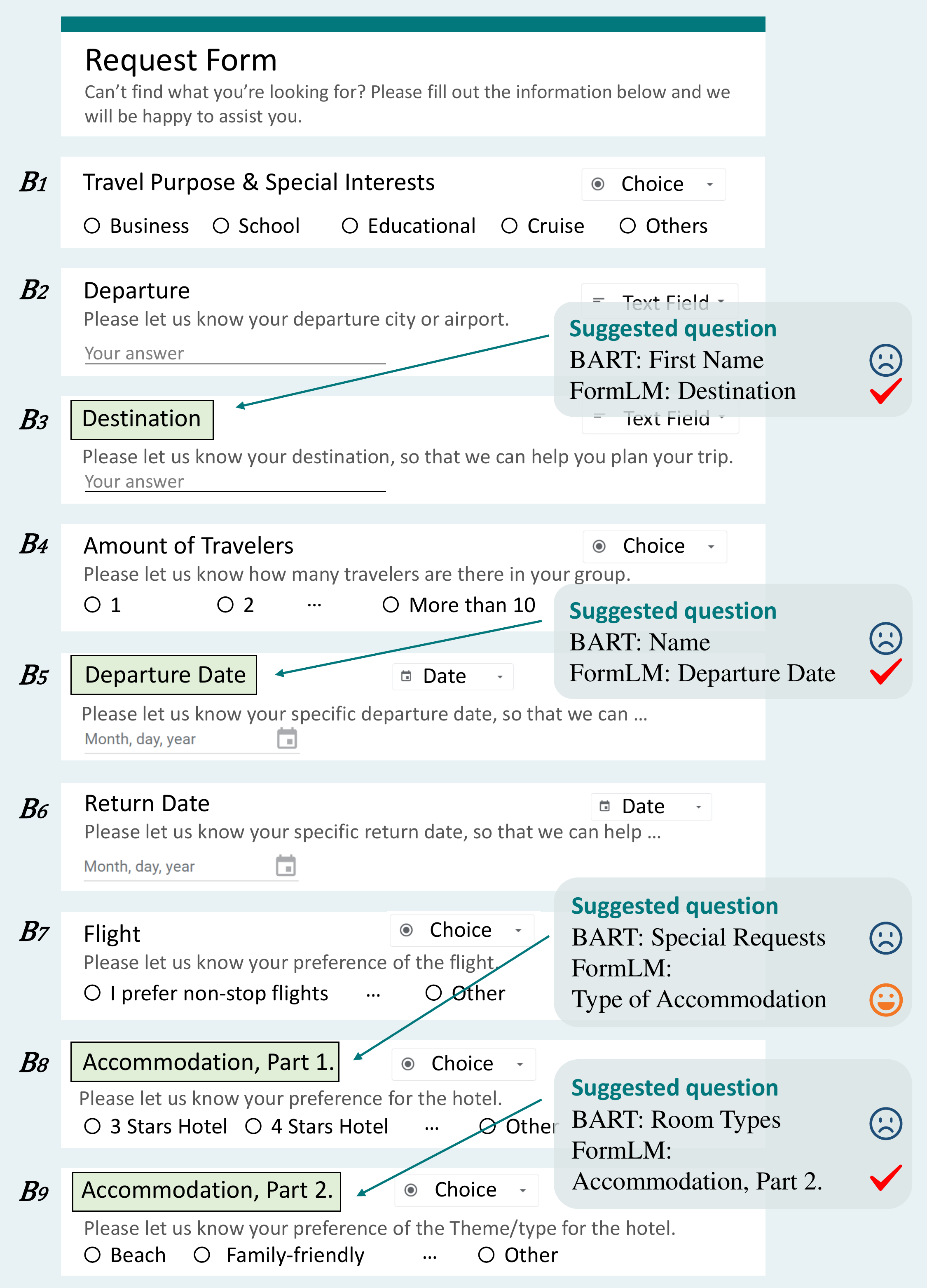}
    \caption{Sample Outputs by FormLM and BART for Question Recommendation. FormLM's recommended questions are more relevant to the topic and more suitable to the selected block type.}
    \label{Fig.case_study}
\end{figure}

Compared to its backbone BART model (well-known for conditional generation tasks), FormLM further improves the ROUGE-1 scores by 4.71 and 1.12 on Question and Options Recommendations. Human evaluation results in Appendix~\ref{sec:human} also confirm the superiority of FormLM over other baseline models in these two generation tasks. Figure~\ref{Fig.case_study} shows questions recommended by BART and FormLM on an example form from the test set. FormLM's recommendations (\eg, ``Destination'', ``Departure Date'') are more specific and more relevant to the topic of this form, while BART's recommendations (\eg, ``Name'', Special Requests'') are rather general. 
Also, after users create $B_1,B_2,B_3,B_4$ and select $B_5$ as a \textit{Date} type block, FormLM recommends ``Departure Date'' while BART recommends ``Name'' which is obviously not suitable to $B_5$. 

On Block Type Suggestion, FormLM improves the Macro-F1 score by 10.6. The improvement of FormLM over BART ($\uparrow$ rows in Table~\ref{table:mainResults}) shows that our method is highly effective. We will further analyze this in \cref{sec:ablation}.



Note that MarkupLM is a very strong baseline for Block Type Suggestion. This model can partly capture the structural information by parsing the form as a DOM~\citep{wood1998document} 
tree. 
However, since MarkupLM is not specifically designed for online forms, 
it is still 4.1 points worse in Macro-F1 than FormLM on this task.

\subsection{Analysis of FormLM Designs}
\label{sec:ablation}

\begin{table}[ht]
\centering
\small
\begin{tabular}{lccc} 
\toprule
                        & \textbf{Question} & \textbf{Options} & \textbf{Type}   \\
                        & R2               & R2                & F1              \\ 
\hline
\textbf{Full Model}     & \textbf{19.70}   & \textbf{34.65}    & \textbf{83.9}  \\
~$-$
  Decoder StructAttn & 18.90            & 34.36             & 83.7           \\
~$-$
  Encoder StructAttn & 19.58            & 34.41             & 77.9            \\
~$-$
  Form Serialization & 17.43            & 33.83             & 75.5            \\
~$-$
  Previous Context   & 12.67            & 27.65             & 71.8            \\
\bottomrule
\end{tabular}
\caption{Ablation Studies on Form Serialization and Structural Attention. ``$-$'' means the corresponding component is sequentially removed from FormLM. 
``$-$ Previous Context'' means that the closest block title is the only input.
}
\label{table:ablation1}
\end{table}

To further investigate the effectiveness of the design choices in FormLM, we conduct ablation studies and controlled experiments (which are fine-tuned under the same settings as described in \cref{sec:setups}) on the following aspects. 

\noindent
\textbf{Form Serialization}\quad
For Form Creation Ideas, it is important to model the complete form context (defined in \cref{sec:problem}).
Row ``$-$ Previous Context'' of \reftab{table:ablation1} shows that there is a large performance drop on all the tasks if block title is the only input.\footnote{For ablation studies in \reftab{table:ablation1}, the components are sequentially removed because StructAttn depends on the tree structure preserved in form serialization and both techniques become meaningless if we don't model the form context.}

Therefore, we also study the effect of form serialization (see \cref{sec:linearization}) which flattens the form context while preserving its tree structure.  
A naive way of serialization is directly concatenating all available text as NL inputs. Results in this setting (row ``$-$ Form Serialization'' of \reftab{table:ablation1}) are much worse than the results of FormLM with form serialization technique. On Block Type Suggestion, the gap is as large as 8.4 on Macro-F1.

\begin{table}
\centering
\small
\begin{tabular}{lccc}
\toprule
              & \textbf{Question}        & \textbf{Options}       & \textbf{Type}            \\ 
        & R2               & R2                & F1              \\ 
\midrule
w/o Type Info & 17.96          & 33.97          & 81.5           \\
w/~ ~Type Info  & \textbf{19.70} & \textbf{34.65} & \textbf{83.9}  \\
\bottomrule
\end{tabular}
\caption{\label{table:type-info}Performance of FormLM ``w/'' and ``w/o'' Incorporating the Block Type Information.}
\end{table}

\noindent
\textbf{Block Type Information}\quad
A unique characteristic of online forms is the existence of block type (see \cref{sec:form structure}). 
To examine whether FormLM can leverage the important block type information, we run a controlled experiment where block type tokens are replaced by with a placeholder token \texttt{<type>} during form serialization (while other tokens are untouched). As shown in \reftab{table:type-info}, removing block type tokens hurts the model performance on all three tasks, which suggests that FormLM can effectively exploit such information.

\noindent
\textbf{Structural Attention}\quad
FormLM enhances its backbone PLM with StructAttn (\cref{sec:attention}). As the row ``$-$ Encoder StructAttn'' of \reftab{table:ablation1} shows, when we ablate StructAttn from FormLM, the Macro-F1 score of Block Type Suggestion drops from 83.9 to 77.9 and the performance on the generation tasks also drops. 
In FormLM, we apply StructAttn to both encoder and decoder parts. We compare it with the setting without modifying the decoder (row ``$-$ Decoder StructAttn'') and find applying StructAttn to both the encoder and decoder yields uniformly better results, which may be due to better alignment between the encoder and decoder.

\begin{figure}
    \centering 
    \includegraphics[width=\columnwidth]{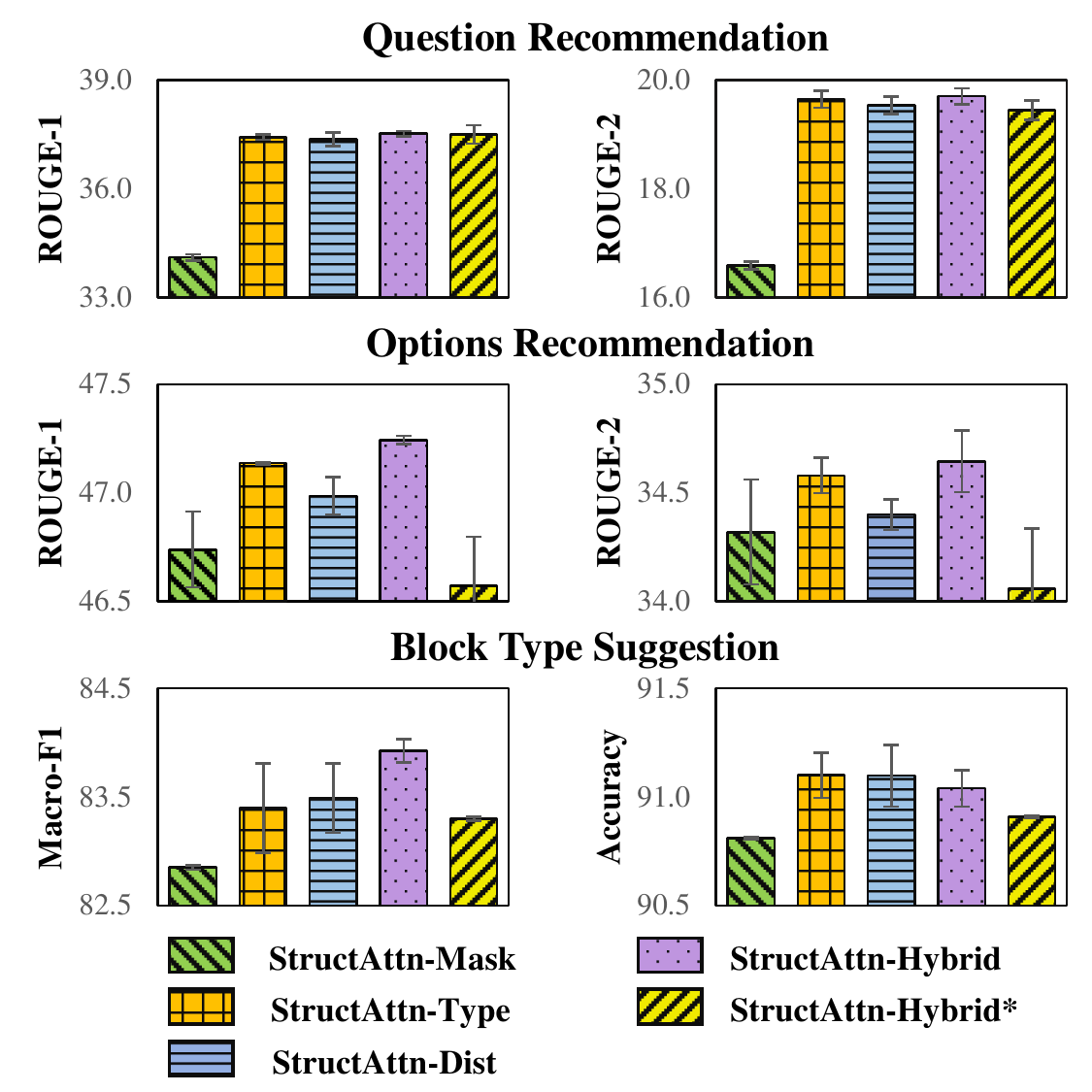}
    \caption{Results of FormLM Using Different Design Choices of StructAttn. (Averaged over 3 runs with std.)}
    \label{Fig.structAttn}
\end{figure}

There are alternative design choices of StructAttn for us to experiment. As \refequ{eq:structAttn} shows, there are two bias terms to model the token type and the block-level distance. We compare this design choice (``Hybrid'' in \reffig{Fig.structAttn}) with adding only the token type bias (``Type'') and only the distance bias (``Dist''). Note that ``Hybrid'' encodes block-level distance through the exponential decay function, we also compare it with another intuitive design (``Hybrid*'') where we use a learnable bias to indicate whether two tokens are within the same block. Besides adding biases, another common practice of modifying attentions is masking. We experiment this design choice (``Mask'') by restricting attentions to those tokens in the same node or parent and grandparent nodes within the tree structure. The comparison results are demonstrated in \reffig{Fig.structAttn}. ``Mask'' performs uniformly worse than adding biases. Among the rest of design choices, ``Hybrid'' shows slightly better performance on Options Recommendation and Block Type Suggestion.

\begin{table}
\small
\centering
\begin{tabular}{lccc} 
\toprule
                 & \textbf{Question} & \textbf{Options} & \textbf{Type}  \\
                 & R2               & R2                & F1             \\ 
\hline
w/o Pre-training & 18.82            & 33.78             & 82.2           \\
BTP              & 19.35            & 34.18             & 83.3           \\
SpanMLM          & 19.42            & 33.94             & 83.3           \\
SpanMLM + BTP    & \textbf{19.70}   & \textbf{34.65}    & \textbf{83.9}  \\
\bottomrule
\end{tabular}
\caption{Ablation Study of Different Continual Pre-training Objectives. (Averaged over 3 runs.)}
\label{table:ablation2}
\end{table}

\noindent
\textbf{Continual Pre-training Objectives}\quad
We design two objectives (\cref{sec:pre-training}), SpanMLM and BTP, to continually pre-train FormLM on OOF dataset for better domain adaptation. \reftab{table:ablation2} shows the ablation results of different objectives. We find FormLM trained with both SpanMLM and BTP performs the best. This suggests SpanMLM which focuses more on the recovery of a single node on the tree and BTP which focuses more on the relationship between different nodes can complement each other.

\section{Related Work}
\label{sec:related_work}
\textbf{(Semi-)Structured Data Modeling}\quad
In this paper, we mainly focus on modeling parsed form data. They follow well-defined structure and are usually created by software such as online services mentioned in \cref{sec:intro}. Existing works~\citep{wang-etal-2022-lilt,xu-etal-2021-layoutlmv2,li-etal-2021-structurallm,appalaraju2021docformer,aggarwal-etal-2020-form2seq, he2017multi} focus on another type of forms, scanned forms (\eg, photos and scanned PDF files of receipts or surveys), and process multi-modal inputs (text, image). These forms requires digitization and parsing before passing to any downstream tasks, which are very different from forms studied in this paper.


To the best of our knowledge, the modelling of parsed forms has not been studied before. 
Existing (semi-)structured data modelling works mainly focus on tables~\citep{yin-etal-2020-tabert,wang2021tuta}, documents~\citep{wan-etal-2021-structure,liu-lapata-2019-hierarchical,10.1145/3342558.3345394}, web pages~\citep{wang2022webformer}, \etc Some works represent the (semi-)structured data as a graph and use graph neural network (GNN) for structural encoding~\citep{wang-etal-2020-rat,cai2021sadga}. Some other works convert (semi-)structured data into NL inputs to directly use PLMs~\citep{gong-etal-2020-tablegpt} or modify a certain part of transformer models -- \eg, embedding layers~\citep{herzig-etal-2020-tapas}, attention layers~\citep{eisenschlos-etal-2021-mate, yang-etal-2022-tableformer}, the encoder architecture~\citep{iida-etal-2021-tabbie}. Although it is possible to convert online forms to HTML pages to use models  like MarkupLM~\citep{li-etal-2022-markuplm}, the results are suboptimal as shown in \cref{sec: main results} because the unique structural information of online forms are not fully utilized.



\noindent
\textbf{Intermediate Pre-training}\quad
In \cref{sec:pre-training} we discussed in FormLM how we adapt a general PLM to the form domain through continual pre-training. Intermediate pre-training of a PLM on the target data (usually in a self-supervised way) has been shown efficient on bridging the gap between PLMs and target tasks~\citep{gururangan-etal-2020-dont,rongali2020continual}. Many domain specific models~\citep{xu-etal-2019-bert,chakrabarty-etal-2019-imho,lee2020biobert}, including those for (semi-)structured data~\citep{yin-etal-2020-tabert,liu2022tapex}, are built with this technique. Following the previous approaches, we design form-specific structure-aware training objectives for the continual pre-training process.


\section{Conclusion}
\label{sec:conclusion}
In this paper, we present FormLM for online form modeling. FormLM jointly consider the semantic and structural information by leveraging the PLM and designing form serialization and structural attention. Furthermore, we continually pre-train FormLM on our collected data with structure-aware objectives for better domain adaptation. An extensive set of experiments show that FormLM outperforms baselines on Form Creation Ideas tasks which assist users in the form creation stage.

\section*{Limitations}
\label{sec:limitations}
In this work, we conduct research on online form modeling for the first time. While effective in the proposed tasks of Form Creation Ideas, FormLM has some limitations. First, FormLM is designed to assist form designers by recommending questions / options and suggesting the block type. We believe there are more to explore in recommending creation ideas and we plan to design more tasks for Form Creation Ideas, like recommending a whole block, auto-completion, \etc, to fully exploit FormLM in the form creation stage. Also, since FormLM performs exceptionally well on Block Type Suggestion, it is worthwhile to consider more fine-grained block types. Second, FormLM only models the form content and leaves out the collected responses. Although form content itself is very informative, it is an important research direction to jointly model online forms and their collected responses for they are useful to other stages of the online form life cycle, especially the form analyzing stage. Furthermore, our collected OOF dataset is limited to English forms and doesn't have manual labels. We hope to enlarge our dataset with non-English forms and investigate the possibility of adding supervised labels to this dataset in the future to further facilitate the study of online forms. 


\section*{Ethics Statement}
\noindent
\textbf{Datasets}\quad
In this work, we collect the public OOF dataset for the research community to facilitate future study of online forms. We believe there is no privacy issue related to this dataset. First, the data sources are public available on the Internet, and are anonymously accessible. We complied with the Robots Exclusion Standard during the data collection stage. Second, our dataset only contains form contents and there are no responses or personal information involved. A checklist has been completed at the researchers' institution to ensure the collected dataset does not have ethical issues.

\noindent
\textbf{Risks and Limitations}\quad
Our work proposes FormLM to model online forms and recommend creation ideas to users in the form designing stage. FormLM uses a pre-trained language model, BART, as the backbone. PLMs have a number of ethical concerns in general, like generating biased or discriminative text~\citep{weidinger2021ethical} and involving lots of computing power in pre-training or fine-tuning~\citep{strubell-etal-2019-energy}. The primary risk of our work is that we formulated Question Recommendation and Options Recommendation as generation tasks, but did not include the post-processing of the generated texts in our pipeline. We suggest post-processing the outputs of FormLM to sift out biased or discriminative text before recommending them to the users when applying our technique to online form services. Designing good post-processing technique is also an interesting avenue for future work.

Another limitation we see from an ethical point of view is that we only consider online forms which use English as the primary language. We are trying to collect online forms in other languages and leave it as a future work to provide a multilingual version of FormLM to assist more users in different parts of the world.

\noindent
\textbf{Computational Resources}\quad
The experiments in our paper require computational resources. However, compared with other LMs pretrained from scratch, FormLM inherits the parameters of its backbone and is continually pre-trained with only 50K online forms. 
It takes around 8 hours to complete the continual pre-training with 8 NVIDIA V100 GPUs. Despite this, we recognize that not all researchers have access to this resource level, and these computational resources require energy. Notably, all GPU clusters within our organization are shared, and their carbon footprints are monitored in real-time. Our organization is also consistently upgrading our data centers in order to reduce the energy use.

\bibliography{anthology,custom}
\bibliographystyle{acl_natbib}

\clearpage
\appendix
\section{Details of Open Online Forms Dataset}
\label{appendix:dataset}
\begin{figure}[ht]
    \centering 
    \includegraphics[width=1\columnwidth]{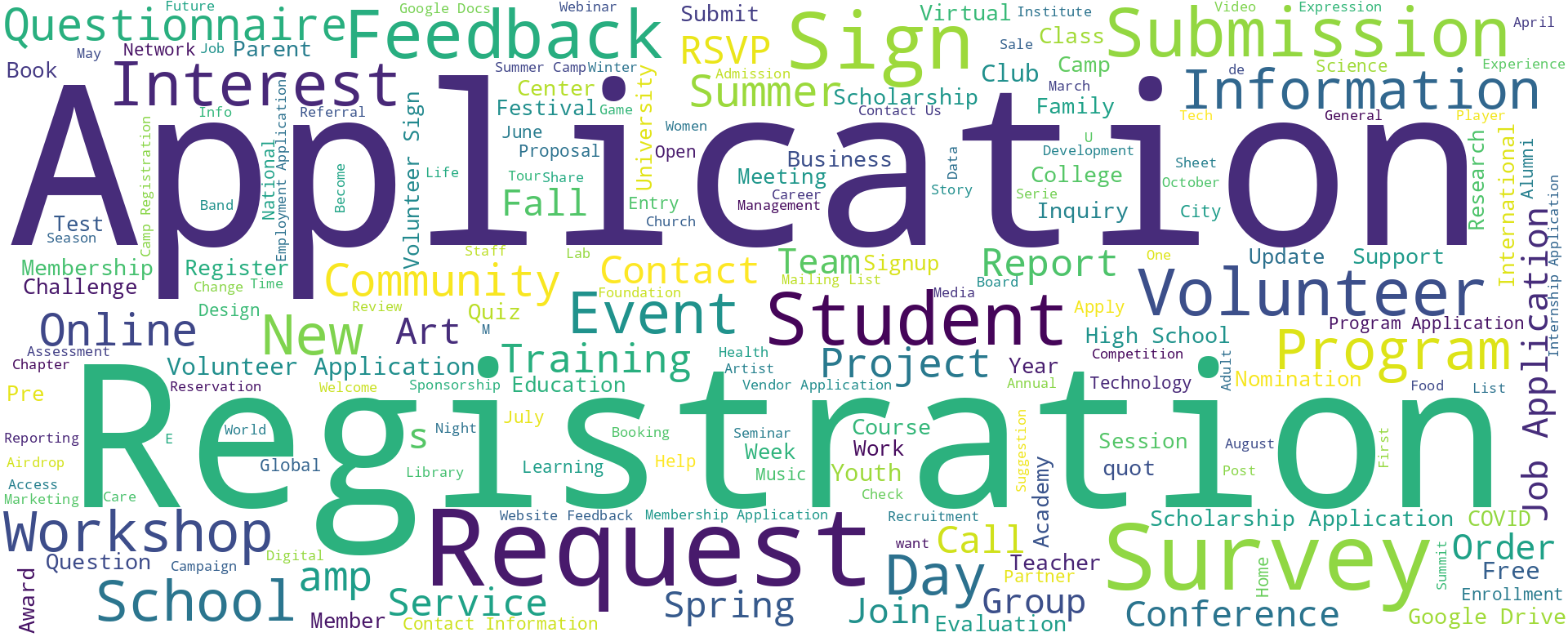}
    \caption{Frequent Words Among Titles of Forms in OOF Dataset.}
    \label{Fig.wc}
\end{figure}

OOF (Open Online Forms) dataset consists of 62K public forms collected on the Web, covering a wide range of domains and purposes. Figure~\ref{Fig.wc} shows some frequent words among titles of the collected data.

\subsection{Dataset Preprocessing}
We crawled 232,758 forms created by a popular online form service on the Internet and filter the crawled data using the following constraints: (1) have at least one question block; (2) have no duplicate question blocks; (3) detected as ``en''\footnote{\url{https://en.wikipedia.org/wiki/List_of_ISO_639-1_codes}} by Language Detection API of Azure Cognitive Service for Language\footnote{\url{https://docs.microsoft.com/en-us/azure/cognitive-services/language-service/language-detection/overview}}. Finally, 62,380 forms meet all constraints. We randomly split them into 49,904 for training, 6,238 for validation and 6,238 for training. 

As introduced in \cref{sec:dataset}, we parsed the crawled HTML pages into JSON format according to the online form structure. Specifically, each JSON file contains keys of  ``title'', ``description'' and ``body'' which correspond to form title ($T$), form description ($Desc$), and an array of blocks ($\{B_1,\cdots,B_n\}$). Each block contains keys of ``title'', ``description'' and ``type''. For \textit{Choice} type blocks and \textit{Rating} type blocks, they further contain the key of ``options''; for \textit{Likert} type blocks, they further contain keys of ``rows'' and ``columns''. For \textit{Description} block, we only keep the plain NL text and remove possible information of other modalities (\textit{i.e}, image, video) because only around 0.1\% of \textit{Description} blocks contain video and 2.0\% contain image. When parsing the HTML pages into JSON format, we also remove non-ASCII characters within the form.


\subsection{Form Length Distribution}

\begin{figure}[h]
    \centering 
    \includegraphics[width=1\columnwidth]{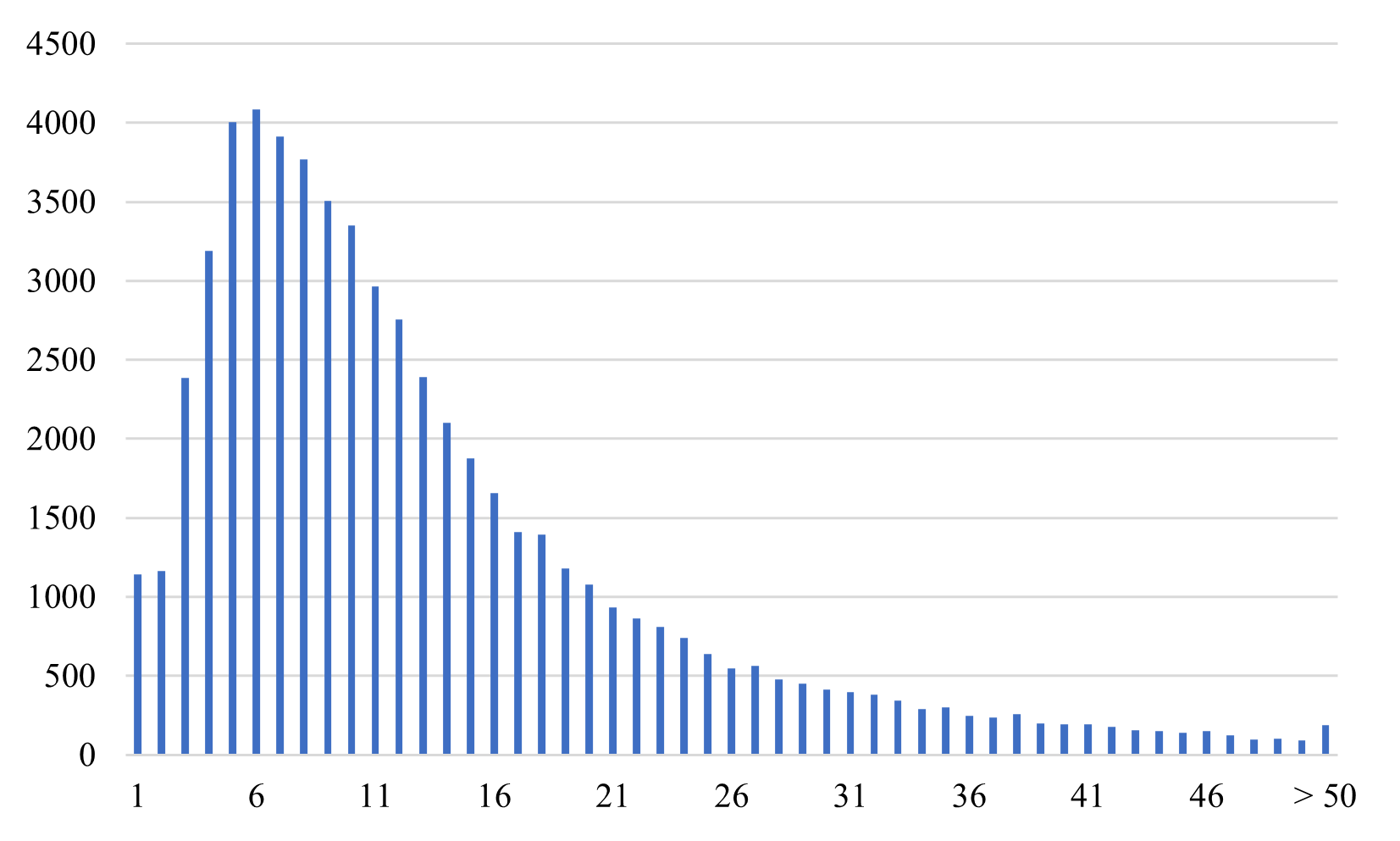}
    \caption{Form Length Distribution of Forms in OOF Dataset.}
    \label{Fig.length}
\end{figure}

We define the length of an online form as the number of blocks within it. Around 80\% of collected forms have a form length no greater than 20. The detailed distribution of form length is shown in \reffig{Fig.length}. As we have discussed in \cref{sec:data_and_metric}, we further perform random sampling to construct our experiment dataset to avoid sample biases introduced by those lengthy forms.

\section{Model Configurations}
\label{sec:config}
We compare FormLM with four baseline models, RoBERTa, GPT-2, MarkupLM, and BART. FormLM adds a small number of additional parameters to its backbone model (278K for FormLM and 208K for $\text{FormLM}_\text{BASE}$) to encode structural information in attention layers (\cref{sec:attention}). \reftab{table:config} shows model configurations of FormLM and baselines in our experiments.

\begin{table}[h]
\centering
\small
\begin{tabular}{lll} 
\toprule
\textbf{Model} & \textbf{\#Params} & \textbf{\#Layers}  \\ 
\midrule
RoBERTa        & 124M              & 12                 \\
GPT-2          & 124M              & 12                 \\
MarkupLM       & 135M              & 12                 \\
$\text{BART}_\text{BASE}$      & 139M              & 6+6                \\
BART           & 406M              & 12+12              \\
$\text{FormLM}_\text{BASE}$    & 139M              & 6+6                \\
FormLM         & 406M              & 12+12              \\
\bottomrule
\end{tabular}
\caption{Model Configurations of FormLM and Baselines.}
\label{table:config}
\end{table}

\section{More Implementation Details}
\label{appendix:implementation}
\noindent
\textbf{Continual Pre-training Details}\quad
We conduct continual pre-training on the training set of the OOF dataset using SpanMLM and BTP objectives (\cref{sec:pre-training}). We adopt a masking budget of 15\% in SpanMLM and do BTP on all training samples. We train FormLM for 15K steps on 8 NVIDIA V100 GPUs with 32G GPU memory. We set the total batch size as 32 and the max sequence length as 512. We use AdamW optimizer~\citep{loshchilov2018decoupled} with $\beta_1=0.9$, $\beta_2=0.999$ and the learning rate of 5e-5. It takes around 8 hours to complete the continual pre-training on our machine.

\noindent
\textbf{Fine-tuning Details}\quad
Among our downstream tasks,  Next Question Recommendation and Options Recommendation are formulated as conditional generation tasks. We use the form serialization procedure (\cref{sec:linearization}) to convert the available context into model inputs. We fine-tune FormLM for 5 epochs with the total batch size of 32, the max source sequence length of 512, and the max target sequence length of 64. We load the best model which has the highest ROUGE-2 score on the validation set in the training process. During generation, we do beam search and set the beam size as 5. Block Type Classification is formulated as a sequence classification task. We follow the original implementation of BART by feeding the same input into the encoder and decoder and passing the final hidden state of the last decoded token into a multi-class linear classifier for classification. We fine-tune FormLM with 5 epochs with the total batch size as 32 and load the best model which has the highest Macro-F1 score on the validation set during the fine-tuning process.

\section{Qualitative Study}
\label{appendix:study}
\begin{figure*}[h]
    \centering 
    \includegraphics[width=0.9\textwidth]{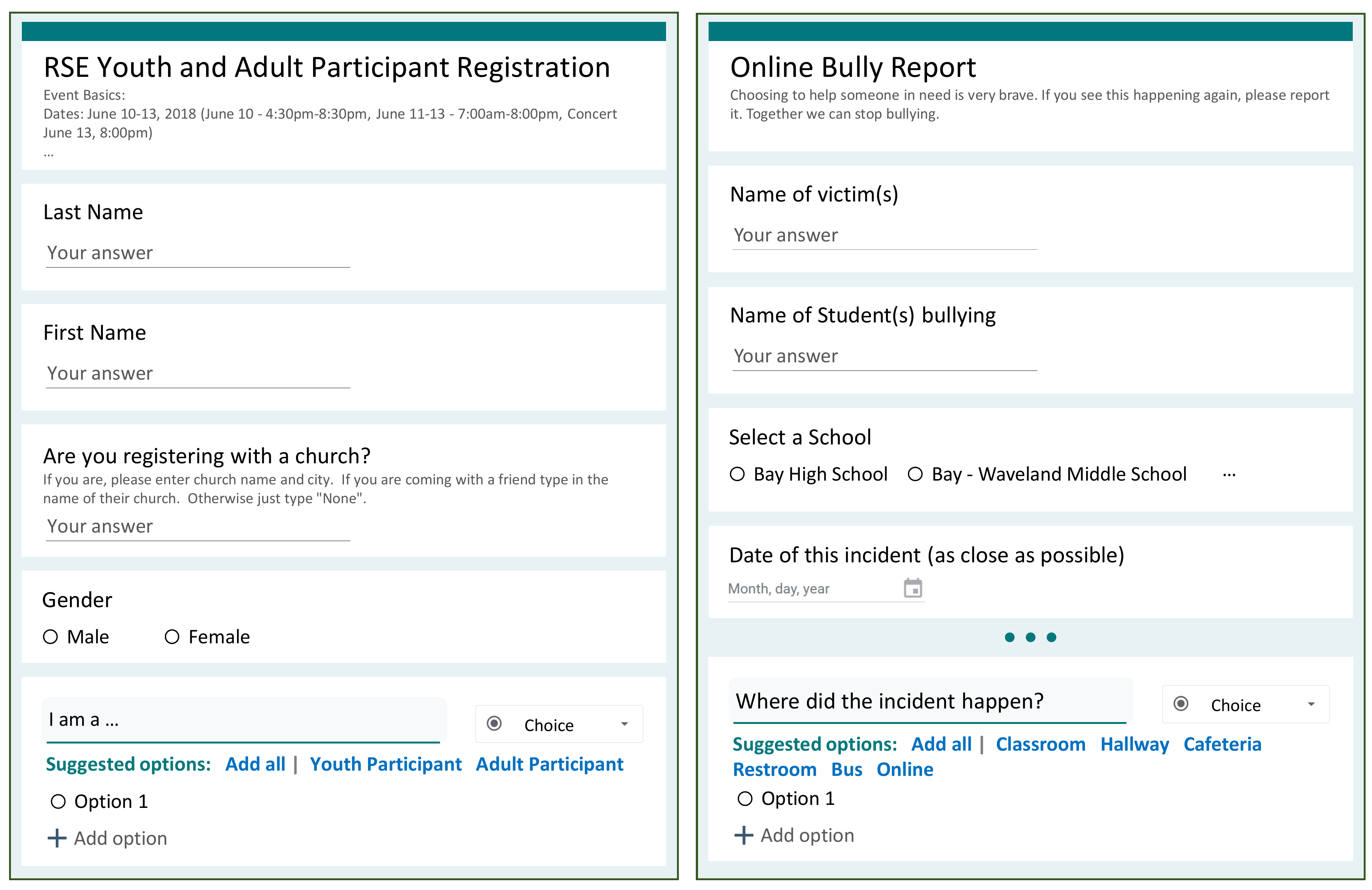}
    \caption{Sample Outputs by FormLM for Options Recommendation. The suggested options are highlighted in \color{myblue}{blue}.}
    \label{Fig.options}
\end{figure*}
Online forms, as a special format of questionnaires, are mainly used to collect information, \ie, demographic information, needs, preferences, \etc~\citep{krosnick2018questionnaire}. As shown in~\reffig{Fig.wc}, the online forms in the OOF dataset are more about objective topics like ``Application'' and ``Registration'' because these information collection scenarios prevail in the daily usage. To collect information effectively, a good questionnaire should 
include questions related to the topic and these questions must be logically connected with each other. Also, for those close-ended questions (the majority of them are \textit{Choice} type questions), they are expected to offer all possible answers for respondents to choose from but not include off-topic options which may cause confusion~\citep{reja2003open}. These criteria of good questionnaires restrict the searching space of online form composition, thus making the automatic recommendation of creation ideas conceptually possible.

In~\cref{sec: main results}, \reffig{Fig.case_study} shows some questions recommended by FormLM. FormLM is able to recommend questions like ``Destination'',  ``Departure Date'', ``Type of Accommodation'' which are highly related to the topic of travelling and can help collect meaningful information for the travel agency. For Options Recommendation, FormLM can accurately identify polar questions and recommend ``Yes'', ``No'' as candidate options. Also, since FormLM is continually pre-trained on a large amount of online forms, it has no difficulty recommending options for those frequently asked questions, \eg, ``Gender'', ``Current Educational Qualifications'', \etc. More interestingly, we notice that FormLM can provide accurate recommendation for questions which are related to their previous contexts. \reffig{Fig.options} gives two sample outputs by FormLM for Options Recommendation. In the left sample, FormLM gives concrete suggestions which are based on the form title; in the right sample, the recommended locations are all related to school, and they accord well with the domain of this form. We assume that such good performance can be attributed to the effective understanding of form structure and context.

\section{Human Evaluation}
\label{sec:human}
\begin{figure*}[h]
    \centering 
    \includegraphics[width=1\textwidth]{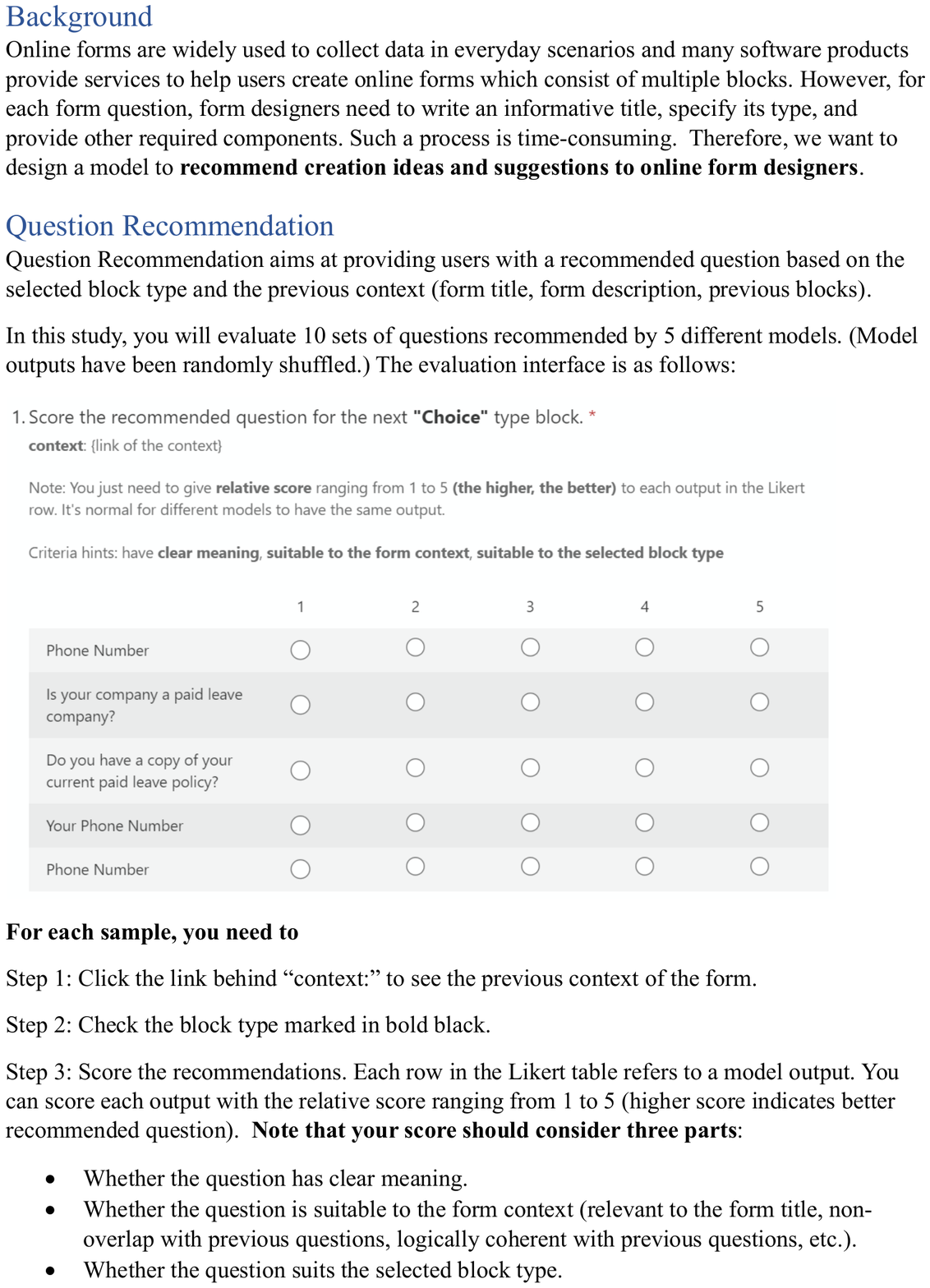}
    \caption{Human Evaluation Instructions. (Page 1 / 2)}
    \label{Fig.eval1}
\end{figure*}

\begin{figure*}[h]
    \centering 
    \includegraphics[width=1\textwidth]{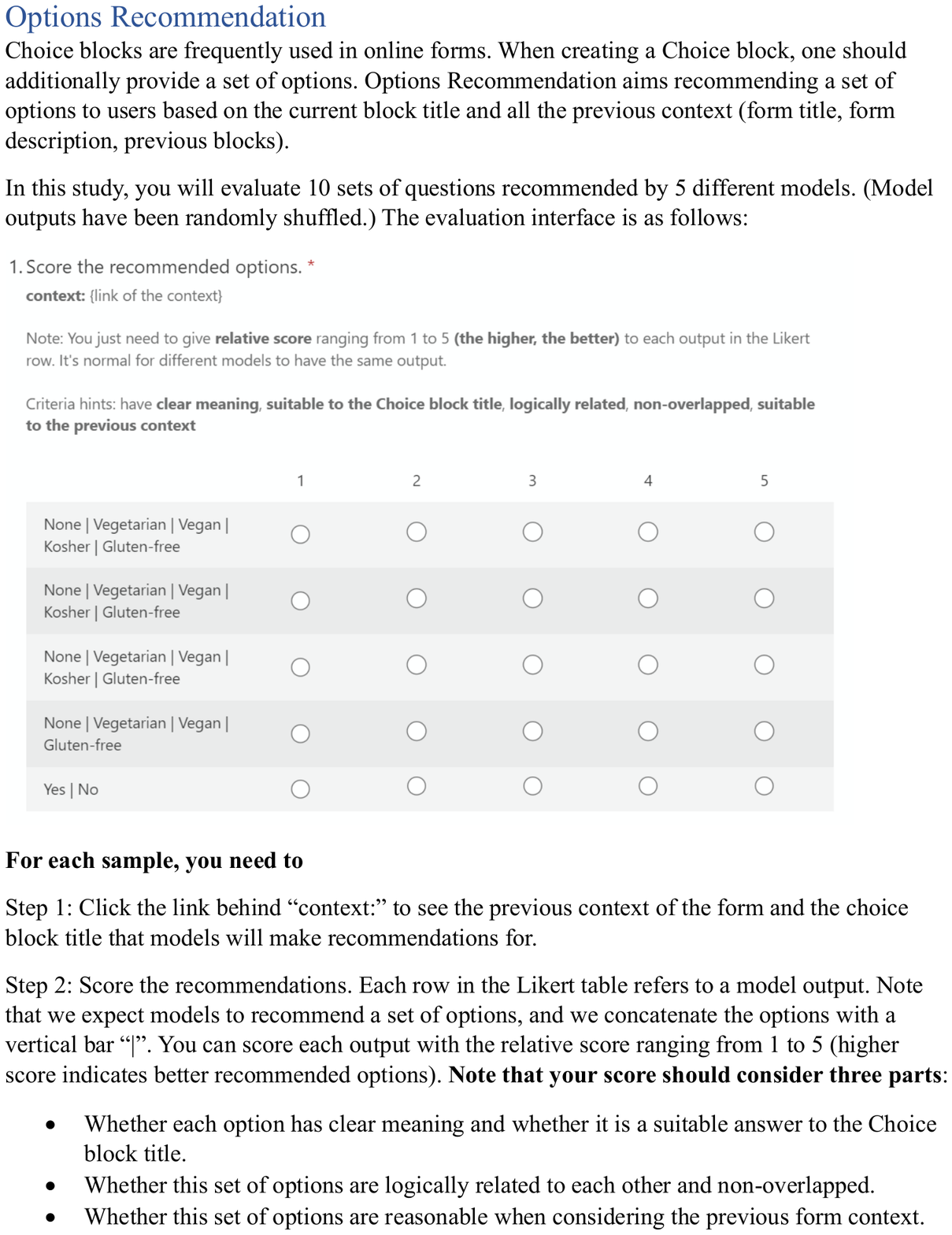}
    \caption{Human Evaluation Instructions. (Page 2 / 2)}
    \label{Fig.eval2}
\end{figure*}

Apart from reporting automatic evaluation results using ROUGE scores, we further conduct human evaluations for Question Recommendation and Options Recommendation. We randomly choose 50 samples from the test sets of the two task and collect the recommended question / options from 5 models (GPT-2, $\text{BART}_\text{BASE}$, BART, $\text{FormLM}_\text{BASE}$, FormLM). We use an HTML website (actually an online form service) to collect the manual labels. Human evaluation instructions are shown in \reffig{Fig.eval1} and \reffig{Fig.eval2}. Eight experts familiar with online form software products participate in the experiment.  For each sample of a task, we construct a Likert question containing the 5 outputs (randomly shuffled and anonymized) of the models. For each sample, three experts compare the 5 outputs using a rating scale of 1 to 5 (the higher, the better) at the same time to achieve better comparison and annotation consistency across different outputs. So in total, we collect 150 expert ratings for each model on each task.

\begin{table}[th]
\centering
\resizebox{\columnwidth}{!}{%
\begin{tabular}{llllllllll} 
\toprule
Rating      & 5  & 4  & 3  & 2  & 1  & Avg. & $\geq$4  & $\geq$3  & $\leq$2  \\ 
\midrule
GPT-2       & 16 & 22 & 23 & 20 & 69 & 2.31 & 38  & 61  & 89  \\
$\text{BART}_\text{BASE}$   & 28 & 21 & 12 & 23 & 66 & 2.48 & 49  & 61  & 89  \\
BART        & 26 & 23 & 25 & 18 & 58 & 2.61 & 49  & 74  & 76  \\
$\text{FormLM}_\text{BASE}$ & 63 & 47 & 13 & 15 & 12 & 3.89 & 110 & 123 & 27  \\
FormLM      & 72 & 41 & 16 & 9  & 12 & \textbf{4.01} & 113 & 129 & 21  \\
\bottomrule
\end{tabular}
}
\caption{Summary of Human Evaluation Ratings for Question Recommendation.}
\label{table:eval_q}
\end{table}

\begin{table}[th]
\centering
\resizebox{\columnwidth}{!}{%
\begin{tabular}{llllllllll} 
\toprule
Rating      & 5  & 4  & 3  & 2  & 1  & Avg. & $\geq$4  & $\geq$3  & $\leq$2  \\ 
\midrule
GPT-2       & 16 & 10 & 6 & 9 & 109 & 1.77 & 26  & 32  & 118  \\
$\text{BART}_\text{BASE}$   & 63 & 28 & 17 & 14 & 28 & 3.56 & 91  & 108  & 42  \\
BART        & 68 & 30 & 23 & 9 & 20 & 3.78 & 98  & 121  & 29  \\
$\text{FormLM}_\text{BASE}$ & 71 & 35 & 18 & 9 & 17 & 3.89 & 106 & 124 & 26  \\
FormLM      & 89 & 29 & 14 & 7  & 11 & \textbf{4.19} & 118 & 132 & 18  \\
\bottomrule
\end{tabular}
}
\caption{Summary of Human Evaluation Ratings for Options Recommendation.}
\label{table:eval_o}
\end{table}

The evaluation results are shown in \reftab{table:eval_q} and \reftab{table:eval_o}. We can see FormLM and $\text{FormLM}_\text{BASE}$ outperform all baseline models on both Question and Options Recommendation when manually evaluated by the experts, which is in accordance with the automatic evaluation results.

We further conduct Wilcoxon signed-rank test~\citep{woolson2007wilcoxon} which is a non-parametric hypothesis test for the matched-pair data to check statistical significance of the comparison between FormLM, $\text{FormLM}_\text{BASE}$ and their backbone models. At 95\% confidence level, when comparing FormLM with BART and comparing $\text{FormLM}_\text{BASE}$ with $\text{BART}_\text{BASE}$, both $p$-values from Wilcoxon test are less than 0.005. These results show that our models have better performance on these two generation tasks than their backbone PLMs which are well-known for conditional generation.

\end{document}